\documentclass[10pt,twocolumn,letterpaper]{article}

\usepackage{cvpr}
\usepackage{times}
\usepackage{epsfig}
\usepackage{graphicx}
\usepackage{amsmath}
\usepackage{amssymb}
\usepackage{multirow}
\usepackage{subfigure}
\usepackage{epstopdf}
\usepackage{float}
\usepackage{caption}
\usepackage{authblk}
\usepackage{balance}

	\let\oldFootnote\footnote
	\newcommand\nextToken\relax
	
	\renewcommand\footnote[1]{%
		\oldFootnote{#1}\futurelet\nextToken\isFootnote}
	
	\newcommand\isFootnote{%
		\ifx\footnote\nextToken\textsuperscript{,}\fi}
	\usepackage[pagebackref=true,breaklinks=true,letterpaper=true,colorlinks,bookmarks=false]{hyperref}
	
	 \cvprfinalcopy % *** Uncomment this line for the final submission
	
	\def\cvprPaperID{357} % *** Enter the CVPR Paper ID here

	\ifcvprfinal\fi
	
\begin{document}
		
		%%%%%%%%% TITLE
		\title{ ReconNet: Non-Iterative Reconstruction of Images from Compressively Sensed Random Measurements}
		
		\author[1,2]{Kuldeep Kulkarni}
		\author[1]{Suhas Lohit}
		\author[1,2]{Pavan Turaga}
		\author[3]{Ronan Kerviche}
		\author[3]{Amit Ashok}
		\affil[1]{School of Electrical, Computer, and Energy Engineering, Arizona State University, Tempe, AZ}
		\affil[2]{School of Arts, Media and Engineering, Arizona State University, Tempe, AZ}
		\affil[3]{College of Optical Sciences, University of Arizona, Tucson, AZ}
		\makeatletter
		\let\@oldmaketitle\@maketitle% Store \@maketitle
		\renewcommand{\@maketitle}{\@oldmaketitle% Update \@maketitle to insert...
			\vspace{-0.2cm} \centering	\includegraphics
			[trim={0 0cm 0cm 0cm},clip,width=0.87\linewidth]
			{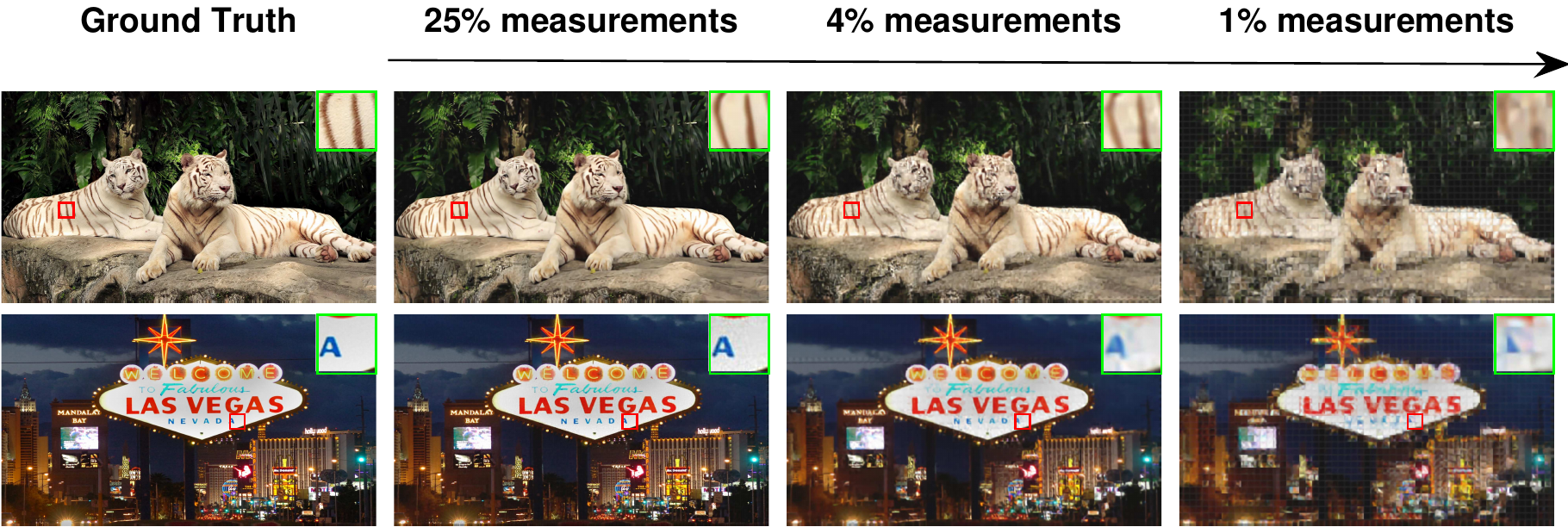} \captionof{figure}{\small{Given the block-wise compressively sensed (CS) measurements, our {\bf non-iterative} algorithm is capable of high quality reconstructions. Notice how fine structures like tiger stripes or letter `A' are recovered from only 4\% measurements. Despite the expected degradation at measurement rate of 1\%, the reconstructions retain rich semantic content in the image. For example, one can easily see that there are two tigers resting on rocks, although the stripes are blurry. This clearly points us to the possibility of CS based imaging becoming a resource-efficient solution in applications, where the final goal is high-level image understanding rather than exact reconstruction.}} \bigskip}% ... an image
		\makeatother
		%\teaser{ \centering \captionsetup{justification=centering} \includegraphics[width=\textwidth]{figures/first_page.eps} \caption{Caption here.} \label{fig:teaser} } 
		
		\maketitle

		%\thispagestyle{empty}
		
		%%%%%%%%% ABSTRACT
		
		\begin{abstract}
			\vspace{-0.4cm}
			The goal of this paper is to present a non-iterative and more importantly an extremely fast algorithm to reconstruct images from compressively sensed (CS) random measurements.  To this end, we propose a novel convolutional neural network (CNN) architecture which takes in CS measurements of an image as input and outputs  an intermediate  reconstruction.  We call this network, ReconNet. The intermediate reconstruction is fed into an off-the-shelf denoiser to obtain the final reconstructed image. % To the best of our knowledge, there exists no published work which proposes either a CNN based or more importantly a non-iterative CS reconstruction algorithm.
			On a standard dataset of  images we show significant improvements in reconstruction results (both in terms of PSNR and time complexity) over state-of-the-art iterative CS reconstruction algorithms at various measurement rates. Further, through qualitative experiments on real data collected using our block single pixel camera (SPC), we show that our network is highly robust to sensor noise and can recover visually better quality images than competitive algorithms at extremely low sensing rates of 0.1 and 0.04. To demonstrate that our algorithm can recover semantically informative images even at a low measurement rate of 0.01, we present a very robust proof of concept real-time visual tracking application. 
		\end{abstract}
		
		%\vspace{-0.5cm}
		%%%%%%%%% BODY TEXT
		\section{Introduction}
		
		%\begin{figure*}
		%\centering
		%\includegraphics[width=0.95\linewidth]{figures/first_page.eps}
		%\caption{Our block-CS image recovery algorithm is capable of high quality reconstructions for diverse set of images.}
		%	\label{fig:recon_0_25_intro}
		%\end{figure*}
		The easy availability of vast amounts of image data and the ever increasing computational power has triggered the resurgence of convolutional neural networks (CNNs) in the past three years and consolidated their position as one of the most powerful machineries in computer vision. 
		%A compelling aspect of CNNs is the seamless manner with which one can adapt them to solve a diverse set of computer vision problems, sometimes even with little or no domain knowledge. 
		Researchers have shown CNNs to break records in the two broad categories of long-standing vision tasks, namely: 1) high-level inference tasks such as image classification , object detection, scene recognition , fine-grained categorization and pose estimation \cite{krizhevsky2012imagenet, girshick2014rich, zhou2014learning, zhang2014part, zhang2014panda} and 2) pixel-wise output tasks like semantic segmentation, depth mapping, surface normal estimation, image super resolution and dense optical flow estimation \cite{Long2015fully, eigen2014depth, Wang2015surface, dong2014learning, walker2015dense}.  
		%Among the per-pixel prediction tasks, of particular interest to us are the tasks of image super resolution and image generation.
		However, the benefits of CNNs have not been explored for one such important task belonging to the latter category, namely reconstruction of images from compressively sensed measurements. In this paper we adapt CNNs to develop an algorithm to recover images from block CS  measurements.  
		
		\vspace{-0.5cm}
		\paragraph*{\textbf{Motivation:}}
		The  advances in compressive sensing theory \cite{donoho2006compressed,candes2006near,Candes} (for the benefit of the readers, a brief background on CS is provided later in the section) has led to the development of many novel imaging devices \cite{SPC,sankaranarayanan2012cs}. The current CS imaging systems, such as the commercially available short-wave infrared single pixel camera, from Inview Technology Corporation, provide the luxury of reduced and fast acquisition of the image by taking only a small number random projections of the scene, thus enabling compression at the sensing level itself.  Such characteristics of the acquisition system are highly sought-after in a) resource-constrained environments like UAVs where generally, computationally expensive methods are employed as a post-acquisition step to compress the fully acquired images, and b) applications such as Magnetic Resonance Imaging (MRI) \cite{lustig2007sparse} where traditional imaging methods are very slow.  As an undesirable consequence, the computational load is now transferred to the decoding algorithm which reconstructs the image from the CS measurements or the random projections. 
		%Barring a few impactful applications such as Magnetic Resonance Imaging (MRI) \cite{lustig2007sparse}, the highly asymmetric nature of the CS theory has made it difficult to put CS imaging into practice.   
		
		Over the past decade,  a plethora of reconstruction algorithms \cite{candes2006robust, duarte2008wavelet, peyre2008non, baraniuk2010model, li2013efficient, kim2010compressed, zhang2013improved,  som2012compressive, metzler2014denoising, dong2014compressive}  have been proposed. However, almost all of them are plagued by a number of  similar drawbacks.  
		Firstly, current approaches solve an optimization problem to reconstruct the images from the CS measurements. Very often, the iterative nature of the solutions to the optimization problems renders the algorithms computationally expensive with some of them even taking as many as 20 minutes to recover just one image, thus making real-time reconstruction impossible. 
		Secondly, in many resource-constrained applications, one may be interested only in some property of the scene like `Where is a particular object in the image?' or `What is the person in the image doing?', rather than the exact values of all pixels in the image. In such scenarios, there is a great urge to acquire as few measurements as possible, and  still be able to recover an image which retains enough information regarding the property of the scene that one is interested in.  The current approaches, although slow, are capable of delivering high quality reconstructions at high measurement rates. However, their performance degrades appreciably as measurement rate decreases, yielding reconstructions which are not useful for any image understanding task. 
		Motivated by these, in this paper we present a CS image recovery algorithm which has the desired features of being computationally light as well as being capable of delivering reasonable quality reconstructions useful for image understanding tasks, even at extremely low  measurement rates of 0.01. The contributions of our paper are the following:
		\vspace{-0.25cm}
		\paragraph*{Contributions:}
		{\bf a)} We propose a {\bf non-iterative} and extremely fast reconstruction algorithm for block CS imaging \cite{gan2007block}.  To the best of our knowledge, there exists no published work which achieves these desirable features.
		{\bf b)} We introduce a novel class of CNN architectures called \textbf{ReconNet} which takes in  CS measurements of an image block as input and outputs the reconstructed image block.  Further, the reconstructed image blocks are arranged appropriately and fed into an off-the-shelf denoiser to recover the full image.
		{\bf c)} Through experiments on a standard dataset of images, we show that, in terms of mean PSNR of reconstructed images, our algorithm beats the nearest competitor by considerable margins at measurement rates of 0.1 and below. Further, we validate the  robustness of ReconNet to arbitrary sensor noise by conducting qualitative experiments on real-data collected using our block SPC. We achieve visually superior quality reconstructions than the traditional CS algorithms. 
		{\bf d)} We demonstrate that the reconstructions retain rich semantic content even at a low measurement rate of 0.01. To this end, we present a proof of concept real-time application, wherein object tracking is performed on-the-fly as the frames are recovered from the CS measurements.
		\vspace{-0.5cm}
		\paragraph*{{Background:}} Compressive Sensing (CS) is a signal acquisition paradigm which provides the ability to sample a signal at sub-Nyquist rates. Unlike traditional sensing methods, in CS, one acquires a small number of random linear measurements, instead of sensing the entire signal, and a reconstruction algorithm is used to recover the original signal from the measurements. Mathematically, the measurements are given by $\mathbf{y} = \Phi \mathbf{x} + \mathbf{e}$, where $\mathbf{x} \in \mathbb{R}^n$ is the signal, $\mathbf{y} \in \mathbb{R}^m$, known as the measurement vector, denotes the set of sensed projections, $\Phi \in \mathbb{R}^{m \times n}$ is called the measurement matrix defined by a set of random patterns, and $\mathbf{e} \in \mathbb{R}^{m}$ is the measurement noise.  Reconstructing $\mathbf{x}$ from $\mathbf{y}$ when $m < n$ is an ill-posed problem.  However,  CS theory \cite{donoho2006compressed,candes2006near} states that the signal $\mathbf{x}$ can be recovered perfectly from a small number of $m$ $=$ $\mathcal{O}$($s$ log($\frac{n}{s}$)) random linear measurements by solving the optimization problem in Eq. \ref{eq:l1}, provided the signal is $s$-sparse in some sparsifying domain,  $\Psi$. 
		\vspace{-0.15cm}
		\begin{equation}\label{eq:l1}
		\min_{\mathbf{x}} \quad||\mathbf{\Psi} \mathbf{x}||_1  \quad \quad s.t \quad \quad  ||\mathbf{y} - {\bf \Phi} \mathbf{x}||_2 \le \epsilon.
		\end{equation}
		
		Variants of the optimization problem with relaxed sparsity assumption in Eq. \ref{eq:l1} have been proposed for the compressible signals as well. However, all such algorithms suffer from drawbacks as already discussed. 
		
		\begin{figure*}[ht]
			\centering
			\vspace{-0.5cm}
			\includegraphics[height=35em, width=1.05\linewidth]{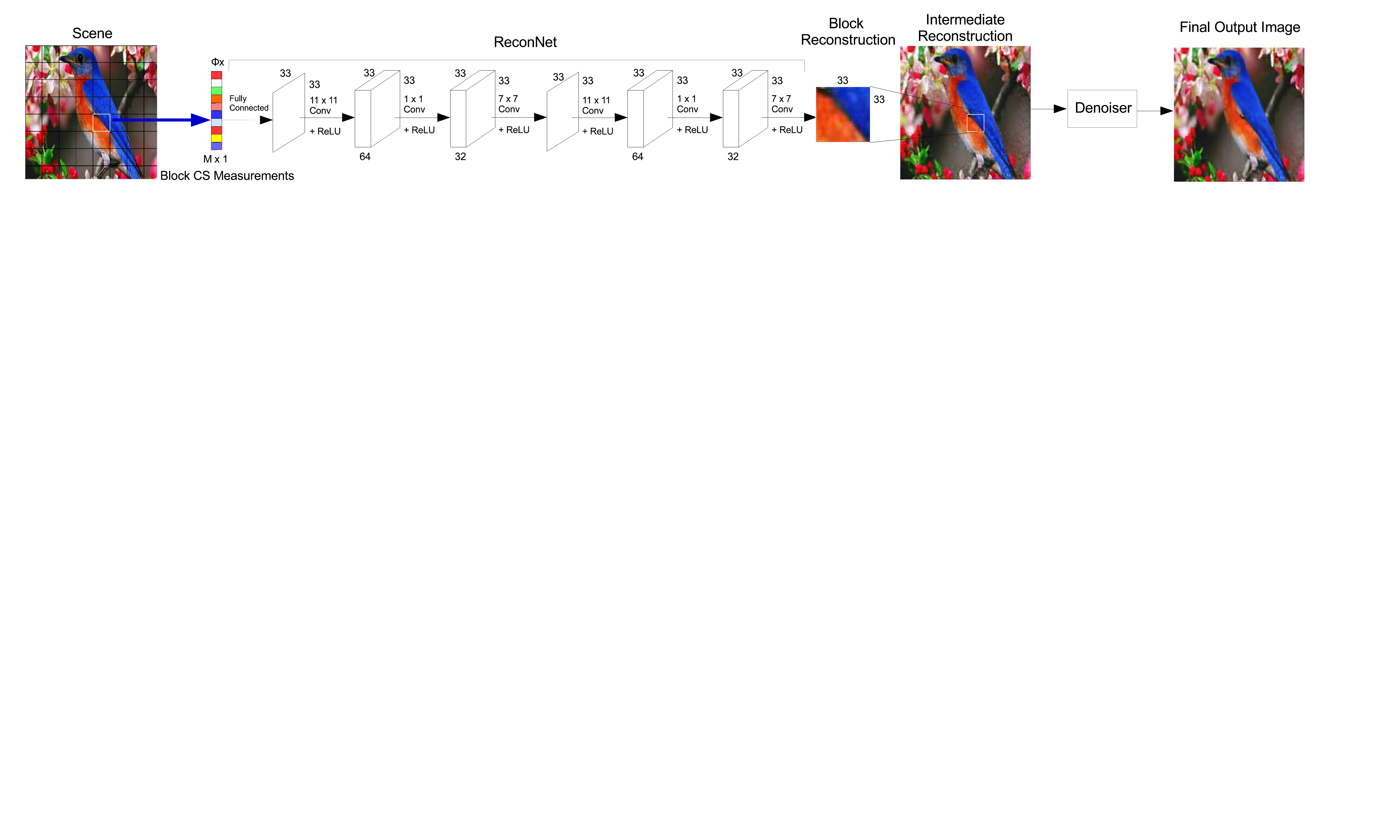}
			\vspace{-10cm}
			\caption{\small{Overview of our non-iterative block CS image recovery algorithm.}}
			\vspace{-0.3cm}
			\label{fig:cscnn_cascade}
		\end{figure*}
		
		\vspace{-0.25cm}
		
		\section{Related Work} \label{sec:related}
		The previous works can be divided into two broad categories, namely CS image reconstruction algorithms and CNNs for per-pixel output tasks.
		\vspace{-0.3cm}
		\paragraph*{CS image reconstruction:} Several algorithms have been proposed to reconstruct images from CS measurements.  The earliest algorithms leveraged the traditional CS theory described above \cite{donoho2006compressed,candes2006near, candes2006robust} and solved the $l_1$-minimization in Eq. \ref{eq:l1} with the assumption that the image is sparse in some transform-domain like wavelet, DCT, or gradient. However,  such sparsity-based algorithms did not work well, since images, though compressible, are not exactly sparse in the transform domain.  This heralded an era of model-based CS recovery methods, where in more complex image models that go beyond simple sparsity were proposed. Model-based CS recovery methods come in two flavors.  In the first, the image model is enforced explicitly  \cite{duarte2008wavelet,  baraniuk2010model, kim2010compressed, som2012compressive}, wherein in each iteration the image estimate is projected onto the solution set defined by the model. These models, often considered under the class of `structured-sparsity' models, are capable of capturing the higher order dependencies between the wavelet coefficients. However, generally a computationally expensive optimization is solved to obtain the projection.
		In the second, the algorithms enforce the image model implicitly through a non-local regularization term in the objective function \cite{peyre2008non,  zhang2013improved,  dong2014compressive}. Recently, a new class of recovery methods called approximate message passing (AMP) algorithms \cite{donoho2009message, tan2015compressive, metzler2014denoising} have been proposed, wherein the image estimate is refined in each iteration using an off-the-shelf denoiser. To the best of our knowledge there exists no published work which proposes a non-iterative solution to the CS image recovery problem.  
		However, there has been one concurrent and independent investigation (paper on arXiv.org, but not yet peer-reviewed or published \cite{mousavi2015deep}) that presents stacked denoising auto-encoders (SDAs) based non-iterative approach for this problem. Different from this, in this paper we present a convolutional architecture, which has fewer parameters, and is easily scalable to larger block-size at the sensing stage, and also results in better performance than SDAs.
		
		\vspace{-0.4cm}
		\paragraph*{CNNs for per-pixel prediction tasks:}  Computer vision researchers have applied CNNs to per-pixel output tasks like semantic segmentation \cite{Long2015fully}, depth estimation \cite{eigen2014depth}, surface normal estimation \cite{Wang2015surface}, image super-resolution \cite{dong2014learning} and dense optical flow estimation from a single image\cite{walker2015dense}. However, these tasks differ fundamentally from the one tackled in this paper in that they map a full-blown image to a similar-sized feature output, while in the CS reconstruction problem, one is required to map a small number of random linear measurements of an image to its estimate. Hence, we cannot use any of the standard CNN architectures that have been proposed so far. Motivated by this, we introduce a novel class of CNN architectures for the CS recovery problem at any arbitrary measurement rate.                     
		
		\vspace{-0.20cm}
		\section{Overview of Our Algorithm} \label{sec:arch}
		Unlike most computer vision tasks like recognition or segmentation to which CNNs have been successfully applied, in the CS recovery problem, the images are not inputs but rather outputs or labels which we seek to obtain from the networks. Hence, the typical CNN architectures which can map images to rich hierarchical visual features are not applicable to our problem of interest. How does one design a network architecture for the CS recovery problem? To answer this question, one can seek inspiration from the CNN-based approach for image super-resolution proposed in \cite{dong2014learning}. Similar to the character of our problem, the outputs in image super-resolution are images, and the inputs -- lower-resolution images -- are of lower dimension. In \cite{dong2014learning}, initial estimates of the high-resolution images are first obtained from low-resolution input images using bicubic interpolation, and then a 3-layered CNN is trained with the initial estimates as inputs and the ground-truth of the desired outputs as labels. If we were to adapt the same architecture for the CS recovery problem, we will have to first generate the initial estimates of the reconstructions from CS measurements. A straightforward option would be to run one of the several existing CS recovery algorithms and obtain initial estimates. But how many iterations do we need to run to ensure a good initial estimate? Running for too many increases computational load, defeating the very goal of this paper of developing a fast algorithm, but running for too few could lead to extremely poor estimates. 
		
		Due to the aforementioned reasons, we relinquish the idea of obtaining initial estimates of the reconstructions, and instead propose a novel class of CNN architectures called ReconNet which can directly map CS measurements to image blocks. The overview of our ReconNet driven algorithm is given in Figure \ref{fig:cscnn_cascade}. The scene is divided into {\bf non-overlapping} blocks. Each block is reconstructed by feeding in the corresponding CS measurements to `ReconNet'. The reconstructed blocks are arranged appropriately to form an intermediate reconstruction of the image, which is input to an off-the-shelf denoiser to remove blocky artifacts and obtain the final output image.  
		
		\vspace{-0.4cm}
		\paragraph*{Network architecture:}      
		Here, we describe the proposed CNN architecture, `ReconNet' shown as part of  Figure \ref{fig:cscnn_cascade}. The input to the network is an $m$-dimensional vector of compressive measurements, denoted by $\Phi\mathbf{x}$, where $\Phi$ is the measurement operator of size $m \times n$, $m$ is the number of measurements and $\mathbf{x}$ is the vectorized input image block. In our case, we train networks capable of reconstructing blocks of size $33 \times 33$, hence $n = 1089$. This block size is chosen so as to reduce the network complexity and hence, the training time, while ensuring a good reconstruction quality. 
		
		The first layer is a fully connected layer that takes compressive measurements as input and outputs a feature map of size $33 \times 33$. The subsequent layers are all convolutional layers inspired by \cite{dong2014learning}. Except the final convolutional layers, all the other layers use ReLU following convolution. All feature maps produced by all convolutional layers are of size $33 \times 33$, which is equal to the block size. The first and the fourth convolutional layers use kernels of size $11 \times 11$ and generate $64$ feature maps each. The second and the fifth convolutional layers use kernels of size $1 \times 1$ and generate 32 feature maps each. The third and the last convolutional layer use a $7 \times 7$ and generate a single feature map, which, in the case of the last layer, is also the output of the network. We use appropriate zero padding to keep the feature map size constant in all layers. 
		%22 135 1400 525
		
		\vspace{-0.4cm}
		\paragraph*{Denoising the intermediate reconstruction:}
		The intermediate reconstruction (see Figure \ref{fig:cscnn_cascade}) is denoised to remove the artifacts resulting due to block-wise processing. We choose BM3D \cite{dabov2007image} as the denoiser since it gives a good trade-off between computational complexity and reconstruction quality. 
		
		\vspace{-0.1cm}
		\section{Learning the ReconNet}
		In this section, we discuss in detail training of deep networks for reconstruction of CS measurements. We use the network architecture shown in Figure \ref{fig:cscnn_cascade} for all the cases.
		
		\vspace{-0.4cm}
		\paragraph{Ground truth for training:} We use the same set of $91$ images as in \cite{dong2014learning}. We uniformly extract patches of size $33 \times 33$ from these images with a stride equal to 14 to form a set of $21760$ patches. We retain only the luminance component of the extracted patches (For RGB images, during test time we use the same network to recover the individual channels). These form the labels of our training set. We obtain the corresponding CS measurements of the patches. These form the inputs of our training set. Experiments indicate that this training set is sufficient to obtain very competitive results compared to existing CS reconstruction algorithms, especially at low measurement rates.
		
		\vspace{-0.4cm}
		\paragraph{Input data for training:} To train our networks, we need CS measurements corresponding to each of the extracted patches. To this end, we simulate noiseless CS as follows. For a given measurement rate, we construct a measurement matrix, $\Phi$ by first generating a random Gaussian matrix of appropriate size, followed by orthonormalizing its rows. Then, we apply $\mathbf{y} = \Phi\mathbf{x}$ to obtain the set of CS measurements, where $\mathbf{x}$ is the vectorized version of the luminance component of an image patch. Thus, an input-label pair in the training set can be represented as ($\Phi\mathbf{x},\mathbf{x}$). We train networks for four different measurement rates (MR) = $0.25, 0.10, 0.04$ and $0.01$. Since, the total number of pixels per block is $n = 1089$, the number of measurements $n = 272, 109, 43$ and $10$ respectively. 
		
		\vspace{-0.3cm}
		\paragraph{Learning algorithm details:} All the networks are trained using Caffe \cite{jia2014caffe}. The loss function is the average reconstruction error over all the training image blocks, given by
		$L(\{W\}) = \frac{1}{T} \sum_i^{T} ||f(\mathbf{y_i},\{W\}) - x_i||^2$, and is minimized by adjusting the weights and biases in the network, $\{W\}$ using backpropagation. $T$ is the total number of image blocks in the training set, $x_i$ is the $i^{th}$ patch and $f(\mathbf{y_i},\{W\})$ is the network output for $i^{th}$ patch. For gradient descent, we set the batch size to $128$ for all the networks. For each measurement rate, we train two networks, one with random Gaussian initialization for the fully connected layer, and one with a deterministic initialization, and choose the network which provides the lower loss on a validation test. For the latter network, the $j^{th}$ weight connecting the $i^{th}$ neuron of the fully connected layer is initialized to be equal to $\Phi^T_{i,j}$. In each case, weights of all convolutional layers are initialized using a random Gaussian with a fixed standard deviation. The learning rate is determined separately for each network using a linear search. All networks are trained on a Nvidia Tesla K40 GPU for about a day each. 
		
		\iffalse
		\begin{figure*}
			\centering
			\fbox{\includegraphics[bb = 140 240 1040 430, width=0.8\linewidth]{figures/training_set.pdf}}
			\caption{A subset of the images used to create the training set. $33 \times 33$ blocks of these images are extracted and used as the ground truth during the training phase.}
			\label{fig:training_set}
		\end{figure*}
		\fi

		\section{Experimental Results}
		In this section, we conduct extensive experiments on both simulated data and real data, and compare the performance of our CS recovery algorithm with state-of-the-art CS image recovery algorithms, both in terms of reconstruction quality and time complexity. 
		\begin{table*}
			\footnotesize
			%\resizebox{\textwidth}{!}{%
			\centering
			\begin{tabular}{|c|c|c|c|c|c|c|c|c|c|}
				\hline
				\multirow{2}{*}{Image Name} & 
				\multirow{2}{*}{Algorithm} & 
				\multicolumn{2}{c|}{MR = 0.25} & 
				\multicolumn{2}{c|}{MR = 0.10} & 
				\multicolumn{2}{c|}{MR = 0.04} & 
				\multicolumn{2}{c|}{MR = 0.01} \\
				\cline{3-10}
				& & w/o BM3D & w/ BM3D & w/o BM3D & w/ BM3D & w/o BM3D & w/ BM3D & w/o BM3D & w/ BM3D\\
				\hline
				% 		\multirow{5}{*}{Monarch} & TVAL3 \cite{li2013efficient} & \textbf{27.77} & \textbf{27.77} & \textbf{21.16} & 21.16 & 16.73 & 16.73 & 11.09 & 11.11\\
				% 		& NLR-CS \cite{dong2014compressive} & 25.91 & 26.06 & 14.59 & 14.67 & 11.62 & 11.97 & 6.38 & 6.71\\
				% 		& D-AMP \cite{metzler2014denoising} & 26.39 & 26.55 & 19.00 & 19.00 & 14.57 & 14.57 & 6.20 & 6.20\\
				% 		& SDA \cite{mousavi2015deep} & 23.54 & 23.32 & 20.95 & 21.04 & 18.09 & 18.19 & 15.31 & 15.38\\
				% 		& ReconNet (Ours) & 24.31 & 25.06 & 21.10 & \textbf{21.51} & \textbf{18.19} & \textbf{18.32} & \textbf{15.39} & \textbf{15.49}\\
				
				% 		\hline
				
				% 		\multirow{5}{*}{Parrot} & TVAL3 & \textbf{27.17} & \textbf{27.24} & 23.13 & \textbf{23.16} & 18.88 & 18.90 & 11.44 & 11.46\\
				% 		& NLR-CS & 26.53 & 26.72 & 14.14 & 14.16 & 10.59 & 10.92 & 5.11 & 5.44\\
				% 		& D-AMP & 26.86 & 26.99 & 21.64 & 21.64 & 15.78 & 15.78 & 5.09 & 5.09\\
				% 		& SDA & 24.48 & 24.36 & 22.13 & 22.35 & \textbf{20.37} & 20.67 & \textbf{17.70} & 17.88\\
				% 		& ReconNet (Ours)& 25.59 & 26.22 & 22.63 & \textbf{23.23} & 20.27 & \textbf{21.06} & 17.63 & \textbf{18.30}\\
				
				% 		\hline
				
				\multirow{5}{*}{Barbara} & TVAL3 \cite{li2013efficient}& 24.19 & 24.20 & 21.88 & 22.21 & 18.98 & 18.98 & 11.94 & 11.96\\
				& NLR-CS \cite{dong2014compressive}& \textbf{28.01} & \textbf{28.00} & 14.80 & 14.84 & 11.08 & 11.56 & 5.50 & 5.86\\
				& D-AMP \cite{metzler2014denoising}& 25.89 & 25.96 & 21.23 & 21.23 & 16.37 & 16.37 & 5.48 & 5.48\\
				& SDA \cite{mousavi2015deep}& 23.19 & 23.20 & 22.07 & 22.39 & \textbf{20.49} & 20.86 & 18.59 & 18.76\\
				& Ours & 23.25 & 23.52 & \textbf{21.89} & \textbf{22.50} & 20.38 & \textbf{21.02} & \textbf{18.61} & \textbf{19.08}\\
				\hline
				
				% 		\multirow{5}{*}{Boats} & TVAL3 & 28.81 & 28.81 & 23.86 & 23.86 & 19.20 & 19.20 & 11.86 & 11.88\\
				% 		& NLR-CS & 29.11 & \textbf{29.27} & 14.82 & 14.86 & 10.76 & 11.21 & 5.38 & 5.72\\
				% 		& D-AMP & \textbf{29.26} & 29.26 & 21.95 & 21.95 & 16.01 & 16.01 & 5.34 & 5.34\\
				% 		& SDA & 26.56 & 26.25 & 24.03 & \textbf{24.18} & 21.29 & 21.54 & \textbf{18.54} & 18.68\\
				% 		& ReconNet (Ours)& 27.30 & 27.35 & \textbf{24.15} & 24.10 & \textbf{21.36} & \textbf{21.62} & 18.49 & \textbf{18.83}\\
				
				% 		\hline
				
				% 		\multirow{5}{*}{Cameraman} & TVAL3 & \textbf{25.69} & \textbf{25.70} & \textbf{21.91} & \textbf{21.92} & 18.30 & 18.33 & 11.97 & 12.00\\
				% 		& NLR-CS & 24.88 & 24.96 & 14.18 & 14.22 & 11.04 & 11.43 & 5.98 & 6.31\\
				% 		& D-AMP & 24.41 & 24.54 & 20.35 & 20.35 & 15.11 & 15.11 & 5.64 & 5.64\\
				% 		& SDA & 22.77 & 22.64 & 21.15 & 21.30 & \textbf{19.32} & 19.55 & 17.06 & 17.19\\
				% 		& ReconNet (Ours)& 23.15 & 23.59 & 21.28 & 21.66 & 19.26 & \textbf{19.72} & \textbf{17.11} & \textbf{17.49}\\
				
				% 		\hline
				
				\multirow{5}{*}{Fingerprint} & TVAL3 & 22.70 & 22.71 & 18.69 & 18.70 & 16.04 & 16.05 & 10.35 & 10.37\\
				& NLR-CS & 23.52 & 23.52 & 12.81 & 12.83 & 9.66 & 10.10 & 4.85 & 5.18\\
				& D-AMP & 25.17 & 23.87 & 17.15 & 16.88 & 13.82 & 14.00 & 4.66 & 4.73\\
				& SDA & 24.28 & 23.45 & 20.29 & 20.31 & 16.87 & 16.83 & \textbf{14.83} & 14.82\\
				& Ours & \textbf{25.57} & \textbf{25.13} & \textbf{20.75} & \textbf{20.97} & \textbf{16.91} & \textbf{16.96} & 14.82 & \textbf{14.88}\\
				
				\hline
				
				\multirow{5}{*}{Flintstones} & TVAL3 & 24.05 & 24.07 & 18.88 & 18.92 & 14.88 & 14.91 & 9.75 & 9.77\\
				& NLR-CS & 22.43 & 22.56 & 12.18 & 12.21 & 8.96 & 9.29 & 4.45 & 4.77\\
				& D-AMP & \textbf{25.02} & \textbf{24.45} & 16.94 & 16.82 & 12.93 & 13.09 & 4.33 & 4.34\\
				& SDA & 20.88 & 20.21 & 18.40 & 18.21 & 16.19 & 16.18 & 13.90 & 13.95\\
				& Ours& 22.45 & 22.59 & \textbf{18.92} & \textbf{19.18} & \textbf{16.30} & \textbf{16.56} & \textbf{13.96} & \textbf{14.08}\\
				
				\hline
				
				% 		\multirow{5}{*}{Foreman} & TVAL3 & 35.42 & 35.54 & \textbf{28.69} & \textbf{28.74} & 20.63 & 20.65 & 10.97 & 11.01\\
				% 		& NLR-CS & \textbf{35.73} & \textbf{35.90} & 13.54 & 13.56 & 9.06 & 9.44 & 3.91 & 4.25\\
				% 		& D-AMP & 35.45 & 34.04  & 25.51 & 25.58 & 16.27 & 16.78 & 3.84 & 3.83\\
				% 		& SDA & 28.39 & 28.89  & 26.43 & 27.16 & 23.62 & 24.09 & \textbf{20.07} & 20.23\\
				% 		& ReconNet (Ours)& 29.47 & 30.78 & 27.09 & 28.59 & \textbf{23.72} & \textbf{24.60} & 20.04 & \textbf{20.33}\\
				
				% 		\hline
				
				% 		\multirow{5}{*}{House} & TVAL3 & 32.08 & 32.13 & 26.29 & 26.32 & 20.94 & 20.96 & 11.86 & 11.90\\
				% 		& NLR-CS & \textbf{34.19} & \textbf{34.19} & 14.77 & 14.80 & 10.66 & 11.09 & 4.96 & 5.29\\
				% 		& D-AMP & 33.64 & 32.68 & 24.84 & 24.71 & 16.91 & 17.37 & 5.00 & 5.02\\
				% 		& SDA & 27.65 & 27.86 & 25.40 & 26.07 & 22.51 & 22.94 & \textbf{19.45} & \textbf{19.59}\\
				% 		& ReconNet (Ours)& 28.46 & 29.19 & \textbf{26.69} & \textbf{26.66} & \textbf{22.58} & \textbf{23.18} & 19.31 & 19.52\\
				
				% 		\hline
				
				\multirow{5}{*}{Lena} & TVAL3 & 28.67 & 28.71 & \textbf{24.16} & 24.18 & 19.46 & 19.47 & 11.87 & 11.89\\
				& NLR-CS & \textbf{29.39} & \textbf{29.67} & 15.30 & 15.33 & 11.61 & 11.99 & 5.95 & 6.27\\
				& D-AMP & 28.00 & 27.41 & 22.51 & 22.47 & 16.52 & 16.86 & 5.73 & 5.96\\
				& SDA & 25.89 &  25.70 & 23.81 & 24.15 & 21.18 & 21.55 & 17.84 & 17.95\\
				& Ours & 26.54 & 26.53 & 23.83 & \textbf{24.47} & \textbf{21.28} & \textbf{21.82} & \textbf{17.87} & \textbf{18.05}\\
				
				% 		\hline
				
				% 		\multirow{5}{*}{Peppers} & TVAL3 & 29.62 & 29.65 & \textbf{22.64} & 22.65 & 18.21 & 18.22 & 11.35 & 11.36\\
				% 		& NLR-CS & 28.89 & 29.25 & 14.93 & 14.99 & 11.39 & 11.80 & 5.77 & 6.10\\
				% 		& D-AMP & \textbf{29.84} & \textbf{28.58} & 21.39 & 21.37 & 16.13 & 16.46 & 5.79 & 5.85\\
				% 		& SDA & 24.30 & 24.22 & 22.09 & 22.34 & \textbf{19.63} & 19.89 & \textbf{16.93} & \textbf{17.02}\\
				% 		& ReconNet (Ours)& 24.77 & 25.16 & 22.15 & \textbf{22.67} & 19.56 & \textbf{20.00} & 16.82 & 16.96\\
				
				\hline
				
				\multirow{5}{*}{\textbf{Mean PSNR}} & TVAL3 & 27.84 & 27.87 & \textbf{22.84} & 22.86 & 18.39 & 18.40 & 11.31 & 11.34\\  
				& NLR-CS & 28.05 & \textbf{28.19} & 14.19 & 14.22 & 10.58 & 10.98 & 5.30 & 5.62\\
				& D-AMP & \textbf{28.17} & 27.67 & 21.14 & 21.09 & 15.49 & 15.67 & 5.19 & 5.23\\
				& SDA & 24.72 & 24.55 & 22.43 & 22.68 & 19.96 & 20.21 & \textbf{17.29} & 17.40\\
				& Ours & 25.54 & 25.92 & 22.68 & \textbf{23.23} & \textbf{19.99} & \textbf{20.44} & 17.27 & \textbf{17.55}\\
				
				\hline
			\end{tabular}
			\caption{\small{PSNR values in dB for 4 of the test images using different algorithms at different measurement rates. At low measurement rates of 0.1, 0.04 and 0.01, our algorithm yields superior quality reconstructions than the traditional iterative CS reconstruction algorithms, TVAL3, NLR-CS, and D-AMP. It is evident that the reconstructions are very stable for our algorithm with a decrease in mean PSNR of only 8.37 dB as the measurement rate decreases from 0.25 to 0.01, while the smallest corresponding dip in mean PSNR for classical reconstruction algorithms is in the case of TVAL3, which is equal to 16.53 dB.}}
			\vspace{-0.3cm}
			\label{table:psnr_test}
		\end{table*}
		 \vspace{-0.3cm}
		\paragraph*{Baselines:} We compare our algorithm with three iterative CS image reconstruction algorithms,  TVAL3 \cite{li2013efficient}, NLR-CS \cite{dong2014compressive} and D-AMP \cite{metzler2014denoising}. We use the code made available by the respective authors on their websites. Parameters for these algorithms, including the number of iterations, are set to the default values. We use BM3D \cite{dabov2007image} denoiser since it gives a good trade-off between time complexity and reconstruction quality. The code for NLR-CS provided on author's website is implemented only for random Fourier sampling. The algorithm first computes an initial estimate using a DCT or wavelet based CS recovery algorithm, and then solves an optimization problem to get the final estimate. Hence, obtaining a good estimate is critical to the success of the algorithm. However, using the code provided on the author's website, we failed to initialize the reconstruction for random Gaussian measurement matrix. Similar observation was reported by \cite{metzler2014denoising}. Following the procedure outlined in \cite{metzler2014denoising}, the initial image estimate for NLR-CS is obtained by running D-AMP (with BM3D denoiser) for $8$ iterations. Once the initial estimate is obtained, we use the default parameters and obtain the final NLR-CS reconstruction. We also compare with the unpublished concurrent work \cite{mousavi2015deep} which presents a SDA based non-iterative approach to recover from block-wise CS measurements. At the time of writing, the authors had not made either the training set or the pre-trained models publicly available. Here, we compare our algorithm with our own implementation of SDA, and show that our algorithm outperforms the SDA. For fair comparison, we denoise the image estimates recovered by baselines as well. The only parameter to be input to the BM3D algorithm is the estimate of the standard Gaussian noise, $\sigma$. To estimate $\sigma$, we first compute the estimates of the standard Gaussian noise for each block in the intermediate reconstruction given by $\sigma_i = \sqrt{\frac{||y_i - \Phi x_i||^2}{m}}$, and then take the median of these estimates.  %As stated above, there exists no published work which proposes a non-iterative solution to the CS recovery solution. As we were writing this paper, we came across on arxiv, an unpublished concurrent work \cite{mousavi2015deep} which presents a non-iterative approach to recover images from block-wise CS measurements. Similar to the work presented in this paper, a deep network based on stacked denoising auto-encoders (SDAs) is trained to learn a function to map CS measurements to image blocks. At the time of writing, the authors had not made either the training set or the pre-trained models publicly available. Here, we compare our algorithm with our own implementation of SDA. 
		
		\subsection{Simulated data} For our simulated experiments, we use a standard set of $11$ grayscale images, compiled from two sources \footnote{\tiny{\url{http://dsp.rice.edu/software/DAMP-toolbox}}}\footnote{\tiny{\url{http://see.xidian.edu.cn/faculty/wsdong/NLR\textunderscore Exps.htm}}}. We conduct both noiseless and noisy block-CS image reconstruction experiments at four different measurement rates 0.25, 0.1, 0.04 and 0.01.  
		
		\vspace{-0.4cm}
		
		\paragraph*{Reconstruction from noiseless CS measurements:} 
		\begin{figure*}[ht]
			\centering
			{\includegraphics[trim = 0cm 0 0 0,clip, width=0.9\linewidth]{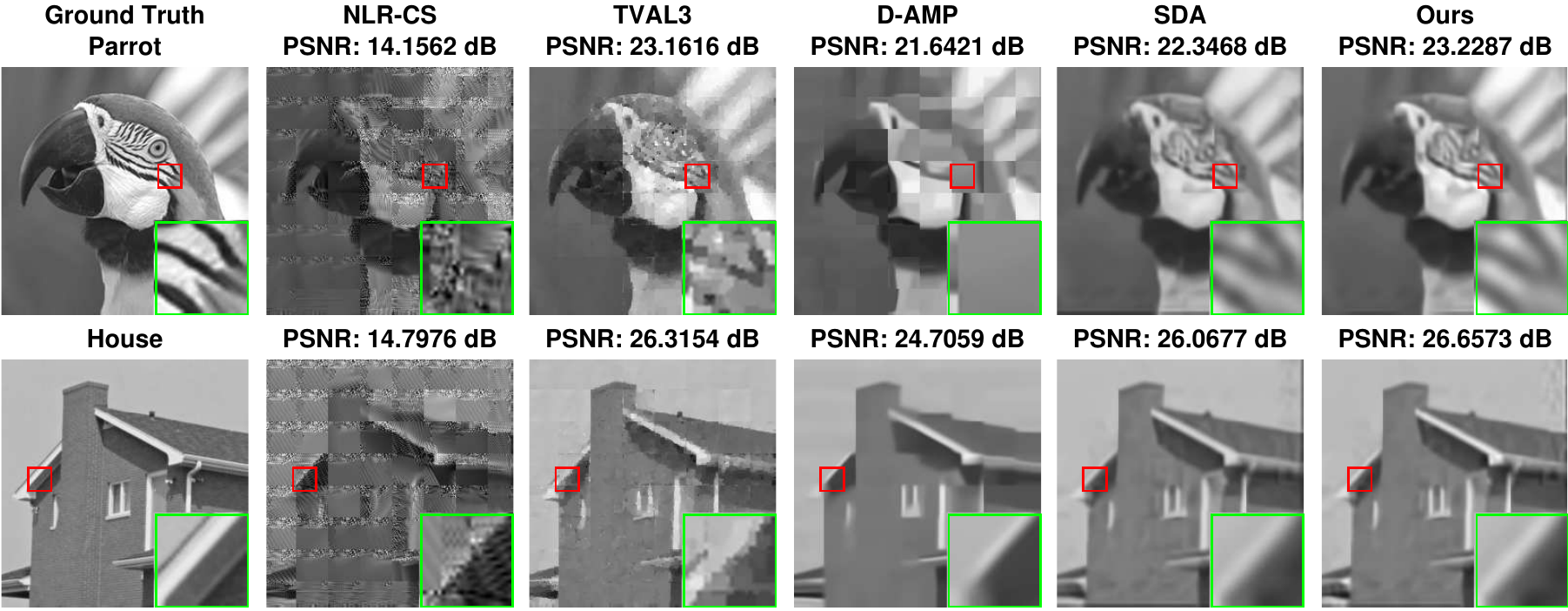}}
			\caption{\small{Reconstruction results for parrot and house images from noiseless CS measurements at measurement rate of 0.1. It is evident that our algorithm recovers more visually appealing images than other competitors. Notice how fine structures are recovered by our algorithm.}}
			\vspace{-0.3cm}
			\label{fig:test_10}
		\end{figure*}
		To simulate noiseless block-wise CS, we first divide the image of interest into non-overlapping blocks of size $33 \times 33$, and then compute CS measurements for each block using the same random Gaussian measurement matrix as was used to generate the training data for the network corresponding to the measurement rate. The PSNR values in dB for both intermediate reconstruction (indicated by w/o BM3D) as well as final denoised versions (indicated by w/ BM3D) for the measurement rates are presented in Table \ref{table:psnr_test}. It is clear from the PSNR values that our algorithm outperforms traditional reconstruction algorithms at low measurement rates of $0.1, 0.04$ and $0.01$. Also, the degradation in performance with lower measurement rates is more graceful. 
		
		Further, in Figure \ref{fig:test_10}, we show the final reconstructions of parrot and house images for various algorithms at measurement rate of 0.1. From the reconstructed images, one can notice that our algorithm, as well as SDA are able to retain the finer features of the images while other algorithms fail to do so. NLR-CS and DAMP provide poor quality reconstruction. Even though TVAL3 yields PSNR values comparable to our algorithm, it introduces undesirable artifacts in the reconstructions. 
		
		\vspace{-0.35cm}
		
		\begin{table}[ht]
			\footnotesize
			%\resizebox{\textwidth}{!}{%
			\centering
			\begin{tabular}{|c|c|c|c|c|c|}
				\hline
				{Algorithm} & {MR = 0.25} & {MR = 0.10} & {MR = 0.04} & {MR = 0.01} \\
				\hline
				TVAL3 & 2.943 & 3.223 & 3.467 & 7.790\\
				\hline
				NLR-CS & 314.852 & 305.703 & 300.666 & 314.176 \\
				\hline
				D-AMP & 27.764 & 31.849 & 34.207 & 54.643\\
				\hline
				ReconNet & 0.0213 & 0.0195 & 0.0192 & 0.0244\\
				\hline
				SDA & 0.0042 & 0.0029 & 0.0025 &  0.0045\\ 
				\hline

			\end{tabular}%}
			\caption{\small{Time complexity (in seconds) of various algorithms (without BM3D) for reconstructing a single $256 \times 256$ image. By taking only about 0.02 seconds at any given measurement rate, ReconNet can recover images from CS measurements in real-time, and is  {\bf $3$} orders of magnitude faster than traditional reconstruction algorithms.}}
			\vspace{-0.85cm}
			\label{table:time}
		\end{table}
		
		\vspace{-0.1cm}
		\paragraph{Time complexity:} In addition to competitive reconstruction quality, for our algorithm without the BM3D denoiser, the computation is real-time and is about {\bf 3} orders of magnitude faster than traditional reconstruction algorithms. To this end, we compare various algorithms in terms of the time taken to produce the intermediate reconstruction of a $256 \times 256$ image from noiseless CS measurements at various measurement rates. For traditional CS algorithms, we use an Intel Xeon E5-1650 CPU to run the implementations provided by the respective authors. For ReconNet and SDA, we use a Nvidia GTX 980 GPU  to compute the reconstructions. The average time taken for the all algorithms of interest are given in table \ref{table:time}. Depending on the measurement rate, the time taken for block-wise reconstruction of a $256 \times 256$ for our algorithm is about $145$ to $390$ times faster than TVAL3, $1400$ to $2700$ times faster than D-AMP, and $15000$ times faster than NLR-CS. It is important to note that the speedup achieved by our algorithm is not solely because of the utilization of the GPU. It is mainly because unlike traditional CS algorithms, our algorithm being CNN based relies on much simpler convolution operations, for which very fast implementations exist. More importantly, the non-iterative nature of our algorithm makes it amenable to parallelization. SDA, also a deep-learning based non-iterative algorithm shows significant speedups over traditional algorithms at all measurement rates.  
		%\begin{tabular}{|*{7}{c|}} 

		\vspace{-0.4cm}
		
\begin{figure*}
			\centering
			{\includegraphics[trim = 0cm 0 0 0,clip, width=0.9\linewidth]{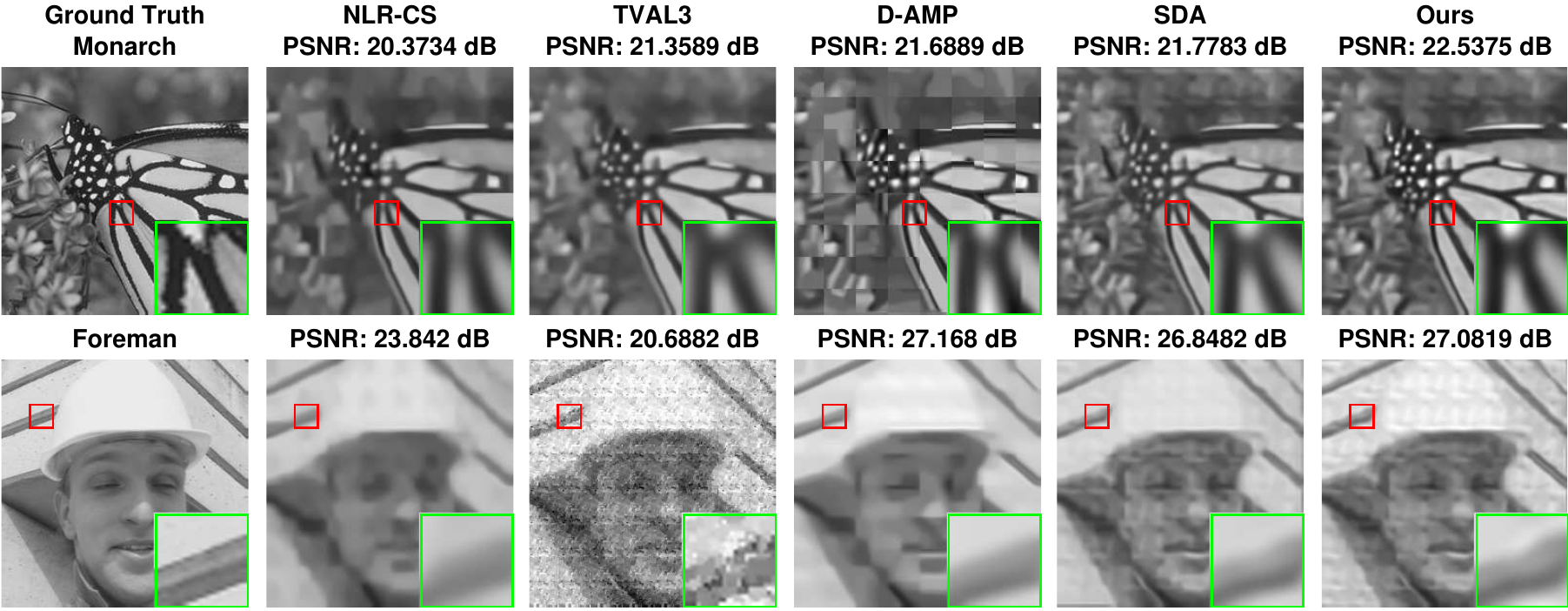}}
			\caption{\small{Reconstruction results for monarch and foreman images from 25\% noisy CS measurements with noise standard deviation equal to 30. One can observe that our algorithm provides visually appealing reconstructions despite high noise level.}}
			\vspace{-0.3cm}
			\label{fig:test_11}
		\end{figure*}

\begin{figure}
			\centering
			\includegraphics[trim = 0cm 0 0 0,clip,width=1\linewidth]{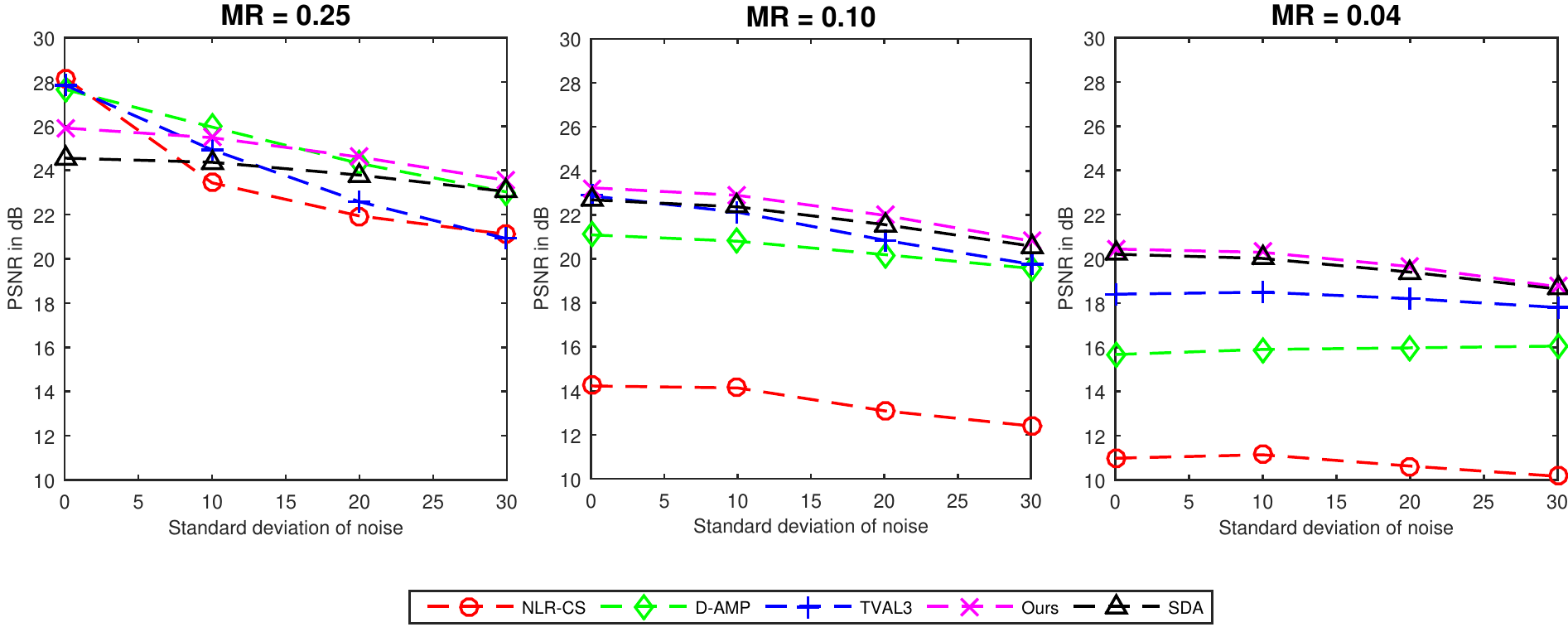}
			\caption{\small{Comparison of different algorithms in terms of mean PSNR (in dB) for the test set in presence of Gaussian noise of different standard deviations at MR = 0.25, 0.10 and 0.04.}}
			\vspace{-0.3cm}
			\label{fig:noise1}
		\end{figure}

		\paragraph*{Performance in the presence of noise:} To demonstrate the robustness of our algorithm to noise, we conduct reconstruction experiments from noisy CS measurements. We perform this experiment at three measurement rates - $0.25, 0.10$ and $0.04$. We emphasize that for ReconNet and SDA,  we {\bf do not} train separate networks for different noise levels but use the same networks as used in the noiseless case. To first obtain the noisy CS measurements, we add standard random Gaussian noise of increasing standard deviation to the noiseless CS measurements of each block. In each case, we test the algorithms at three levels of noise corresponding to $\sigma = 10,20,30$, where $\sigma$ is the standard deviation of the Gaussian noise distribution. The intermediate reconstructions are denoised using BM3D. The mean PSNR for various noise levels for different algorithms at different measurement rates are shown in Figure \ref{fig:noise1}. It can be observed that our algorithm beats all other algorithms at high noise levels. This shows that the method proposed in this paper is extremely robust to all levels of noise. Further, in figure \ref{fig:test_11}, we show the final reconstructions of monarch and foreman images for various algorithms at measurement rate of 0.25 and noise standard deviation equal to $30$. From the reconstructed images, one can notice that our algorithm is extremely robust to noise and provides visually appealing reconstructions despite the very large amount of noise. On the other hand, NLR-CS and TVAL3 provide poor quality reconstruction.  

		\vspace{-0.2cm}
		%\paragraph*{Comparison with SDA \cite{mousavi2015deep}:}
		%From the results presented above, it is clear that ReconNet performs better quality reconstruction results than SDA at almost all measurement rates and noise levels. Apart from raw performance, there are several other advantages ReconNet has over SDA, which we elucidate as below. Firstly, ReconNet is a light-weight network with a total number of parameters equal to 312052, 134545, 62671, and 26734 at measurement rates of 0.25, 0.1, 0.04, and 0.01 respectively, whereas the corresponding numbers of SDA \cite{mousavi2015deep} are 891074, 358390, 142702, and 34858. Secondly, because of the presence of three fully-connected layers, one generally needs to train SDAs for large number of epochs without layer-wise pre-training. However, in comparison `ReconNet' is made up of mainly convolutional layers and just one fully-connected layer. This makes it not only easier to train, but also to scale it to larger networks in terms of both number of layers and the output block size.

		\subsection{Experiments with real data}
		The previous section demonstrated the superiority of our algorithm over traditional algorithms for simulated CS measurements. Here, we show that our networks trained on simulated data can be readily applied for real world scenario by reconstructing images from CS measurements obtained from our block SPC. We compare our reconstruction results with other algorithms. 
		
\begin{figure}[ht]
			\vspace{-0.1cm}
			\centering
			{\includegraphics[trim = 0cm 0 0 0,clip,width=0.9\linewidth]{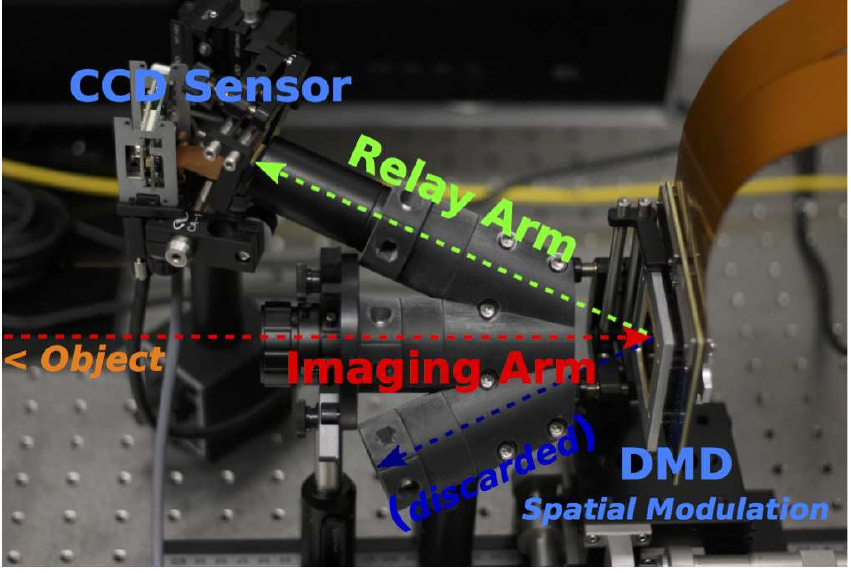}}
			\caption{\small{Compressive imager testbed layout with the object imaging arm in the center, the two DMD imaging arms are on the sides.}}
			\vspace{-0.3cm}
			\label{fig:SystemImage}
		\end{figure}

		\vspace{-0.4cm}
		\paragraph{Scalable Optical Compressive Imager Testbed:}
		We implement a scalable optical compressive imager testbed similar to the one described in \cite{SCIOSA15,kerviche2014information}. It consists of two optical arms and a discrete micro-mirror device (DMD) acting as a spatial light modulator as shown in Figure~\ref{fig:SystemImage}. The first arm, akin to an imaging lens in a traditional system, forms an optical image of the scene in the DMD plane. It has a $40^\circ$ field of view and operates at F/8. The DMD has a resolution of $1920\times1080$ micro-mirror elements, each of size $10.8{\mu}m$. However, in our system the field of view (FoV) is limited to an image circle of 7.5mm, which is approximately 700 DMD pixels. The DMD micro-mirrors are bi-stable and each is either oriented half-way toward the second arm or in the opposite direction (when the flux is discarded). The micro-mirrors can be switched in either direction at a very high rate to effectively achieve 8 bits gray-scale modulation via pulse width modulation. The optically modulated scene on the DMD plane is then imaged (by the second arm) and spatially integrated by a 1/3", $640\times480$ CCD focal plane array with a measurement depth of 12 bits. In the CCD plane, the field of view is 3mm in diameter ($\approx 400$ CCD pixels). Thus, in effect, this testbed implements several single pixel cameras ~\cite{takhar2006new} in parallel. Each block on the DMD effectively maps to a super pixel (e.g. $2 \times 2$ binned pixels) on the CCD. The DMD sequences (in time) through $m$ projections, implementing the $m$ rows of the $m \times n$ projection matrix $\Phi$, where each projection vector appears as a $\sqrt{n} \times \sqrt{n}$ block pattern, replicated across the scene FoV. %Thus for each projection vector $\Phi_i$ we get one compressive measurement $y_i = \Phi_i \mathbf{x} + n_i$ and over $M$ measurements, we get the $M \times 1$ compressed measurement vector $\mathbf{y} = \Phi_i \mathbf{x} + \mathbf{n}$.
		Before data acquisition, a calibration step is performed to map the DMD blocks to CCD detector pixels to characterize any deviation from the idealized system model. 
		
		\vspace{-0.6cm}
		
\begin{figure}[]
			\vspace{-0.1cm}
			\centering
			{\includegraphics[trim = 0cm 0 0 0,clip,width=0.9\linewidth]{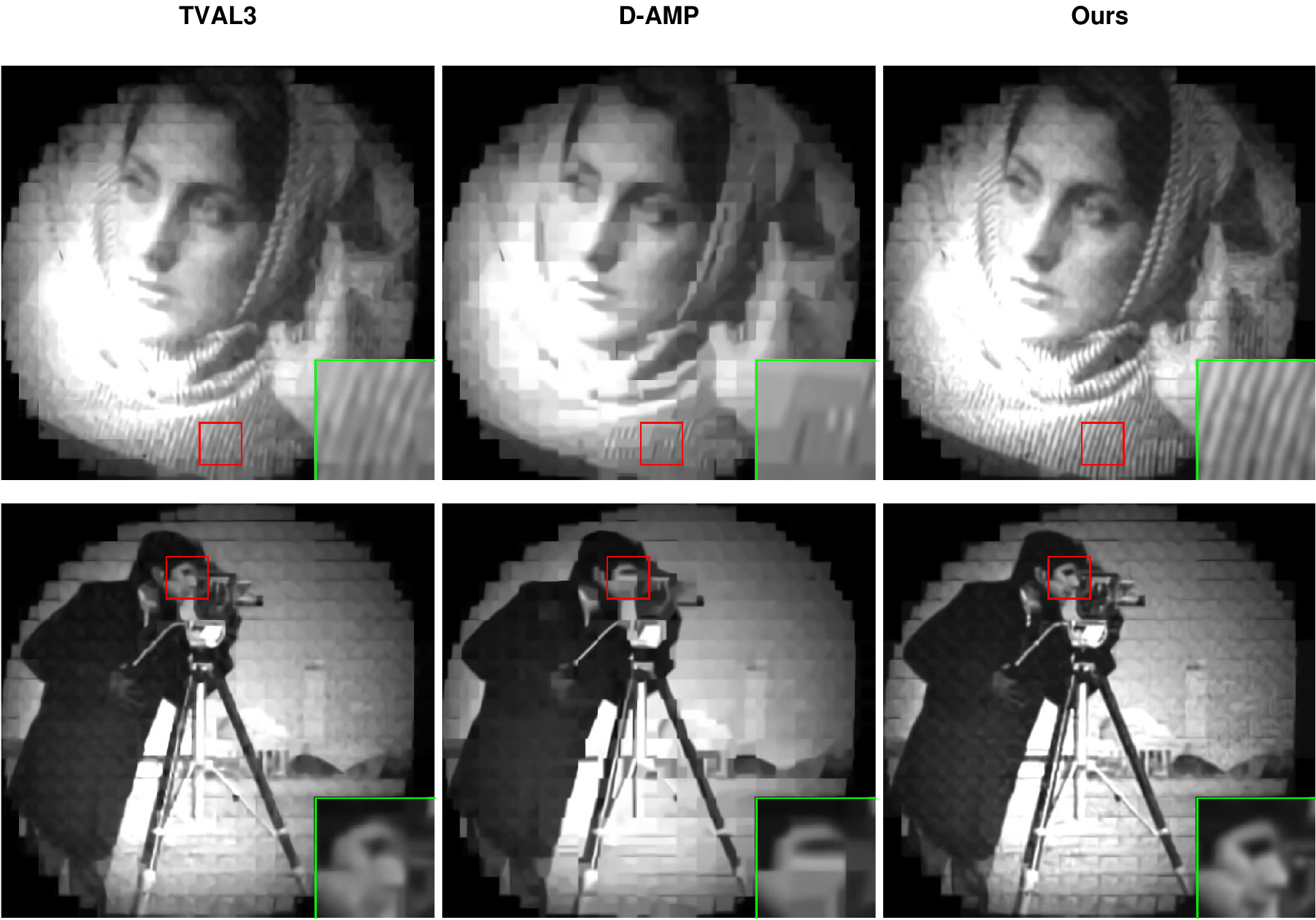}}
			\caption{\small{The figure shows reconstruction results on $3$ images collected using our block SPC operating at measurement rate of 0.1. The reconstructions of our algorithm are qualitatively better than those of TVAL3 and D-AMP. %Notice how finer structures like lines on the clothing or the facial features of the cameraman are recovered using our algorithm. This clearly demonstrates the robustness of our network, and hence also our algorithm to unseen sensor noise.
				}}
			\vspace{-0.3cm}
				\label{fig:real_10}
			\end{figure}

\begin{figure}[]
				\vspace{-0.1cm}
				\centering
				{\includegraphics[trim = 0cm 0 0 0,clip, width=0.9\linewidth]{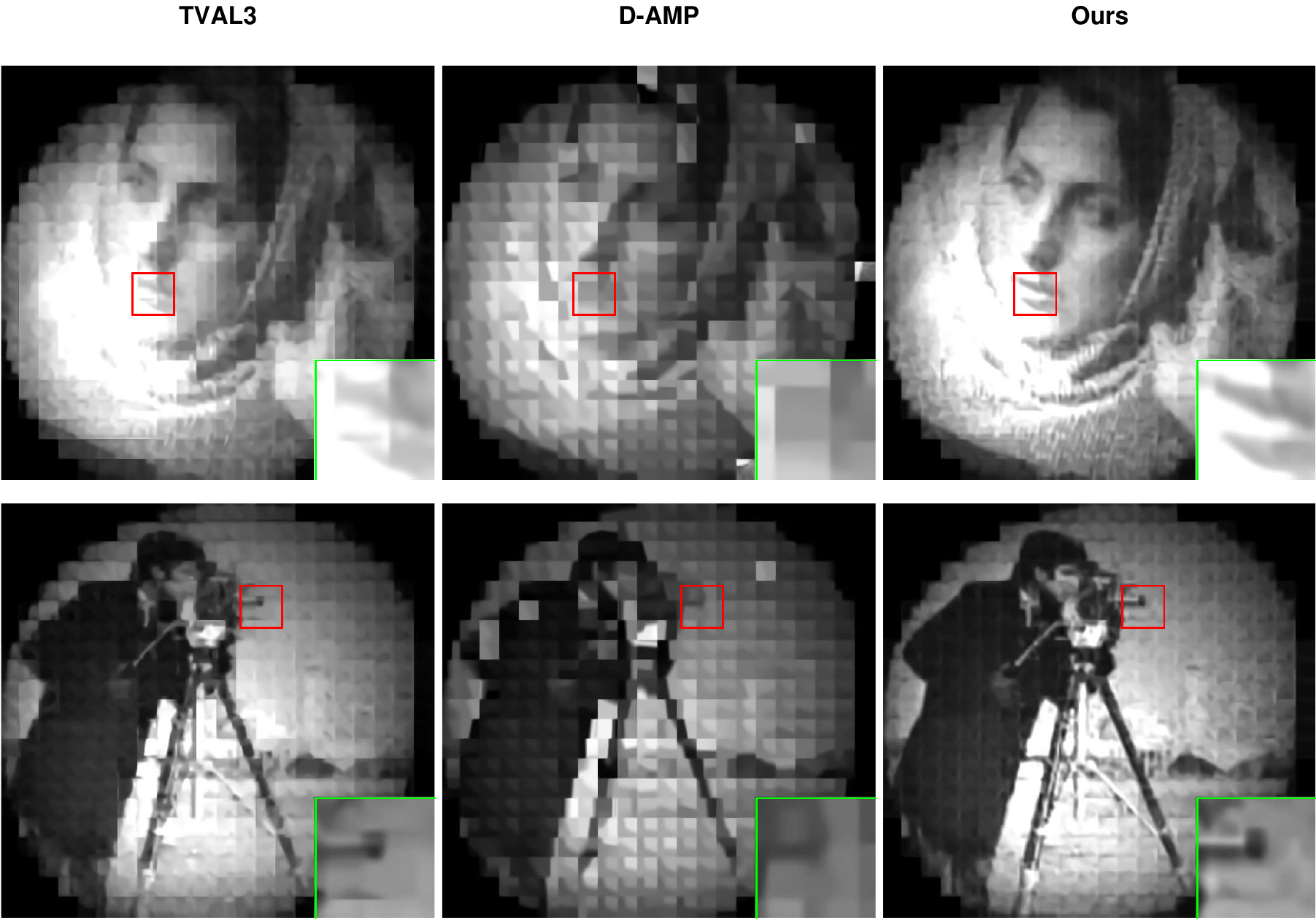}}
				\caption{\small{The figure shows reconstruction results on $3$ images collected using our block SPC operating at measurement rate of 0.04. The reconstructions of our algorithm are qualitatively better than those of TVAL3 and D-AMP. %Despite very low sensing rate of 0.04, our algorithm recovers the facial features like lips and lens of the camera.
					}}
					\vspace{-0.5cm}
					\label{fig:real_04}
				\end{figure}

		\paragraph{Reconstruction experiments:} We use the set up described above to obtain the CS measurements for $383$ blocks (size of $33 \times 33$) of the scene. Operating at MR's of 0.1 and 0.04, we implement the 8-bit quantized versions of measurement matrices (orthogonalized random Gaussian matrices). The measurement vectors are input to the corresponding networks trained on the simulated CS measurements to obtain the block-wise reconstructions as before and the intermediate reconstruction is denoised using BM3D. Figures \ref{fig:real_10} and \ref{fig:real_04} show the reconstruction results using TVAL3, D-AMP and our algorithm for three test images at MR = $0.10$ and $0.04$ respectively. It can be observed that our algorithm yields visually good quality reconstruction and preserves more detail compared to others, thus demonstrating the robustness of our algorithm.
		
				\vspace{-0.2cm}
				
				\section{Real-time high level vision from CS imagers}
\begin{figure}[ht]
					\centering
					{\includegraphics[trim = 0cm 0 0 0,clip, width=0.7\linewidth,height = 11em ]{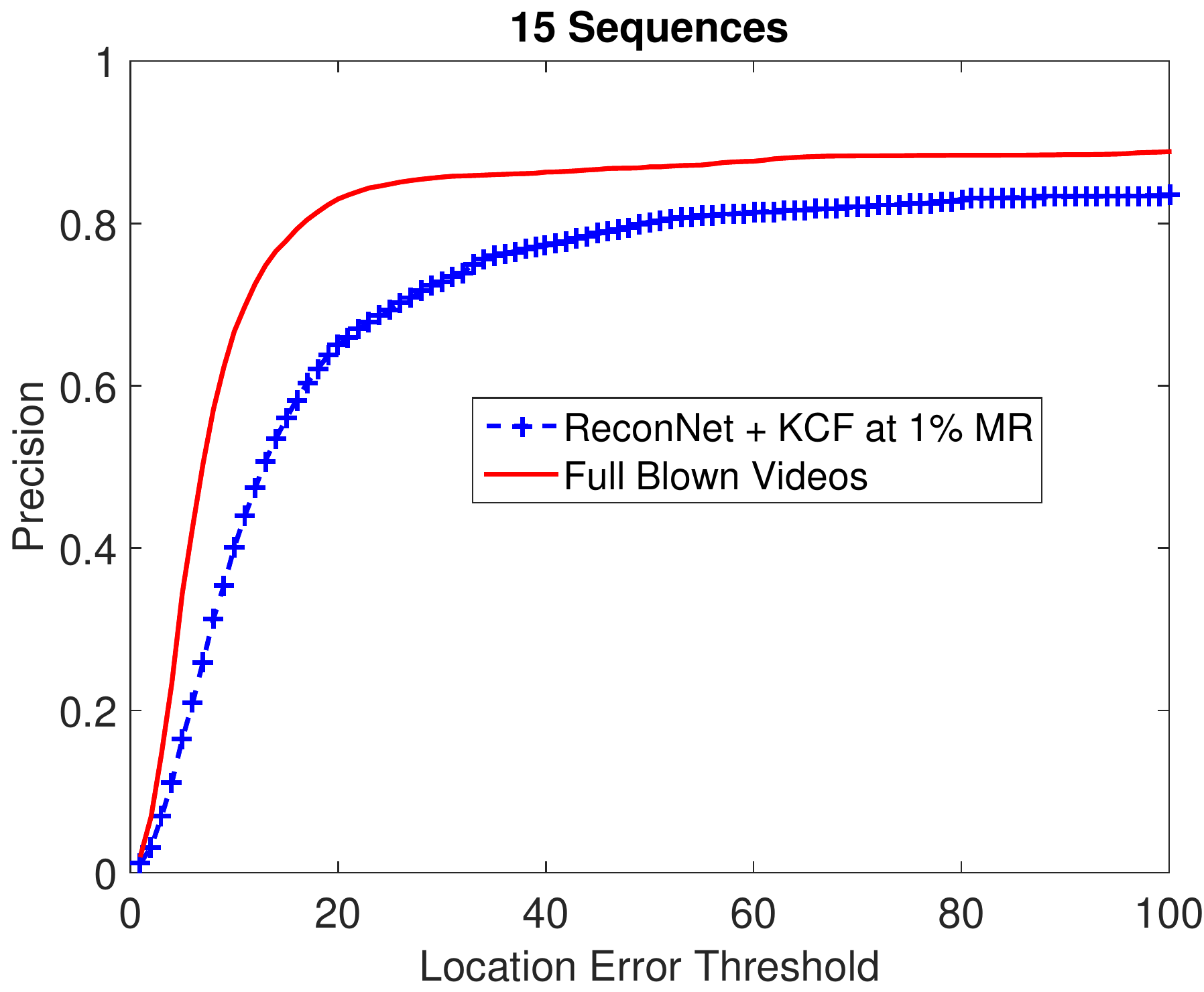}}
					\caption{\small{The figure shows the variation of average precision with location error threshold for ReconNet+KCF and original videos. For a location error threshold of 20 pixels, ReconNet+KCF  achieves an impressive average precision of 65.02\%.}}
					\vspace{-0.3cm}
					\label{fig:tracking}
					
				\end{figure}

				\vspace{-0.2cm}
				In the previous section, we have shown how our approach yields good quality reconstruction results in terms of PSNR over a broad range of measurement rates. Despite the expected degradation in PSNR as the measurement rate plummets to 0.01, our algorithm still yields reconstructions of 15-20 dB PSNR and rich semantic content is still retained. As stated earlier, in many resource-constrained inference applications the goal is to acquire the least amount of data required to perform high-level image understanding. To demonstrate how CS imaging can applied in such scenarios, we present an example proof of concept real-time high level vision application - tracking. To this end we simulate video CS at a measurement rate of 0.01 by obtaining frame-wise block CS measurements on 15 publicly available videos \cite{wu2015object} used to benchmark tracking algorithms. Further, we perform object tracking on-the-fly as we recover the frames of the video using our algorithm without the denoiser. For object tracking we use a state-of-the-art algorithm based on kernelized correlation filters \cite{henriques2015high}. We call the aforementioned pipeline, ReconNet+KCF. For comparison, we conduct tracking on original videos as well. Figure \ref{fig:tracking} shows the average precision curve over the 15 videos, in which each datapoint is the mean percentage of frames that are tracked correctly for a given location error threshold. Using a location error threshold of 20 pixels, the average precision over 15 videos for ReconNet+KCF at 1\% MR is 65.02\%, whereas tracking on the original videos yields an average precision value of 83.01\%. ReconNet+KCF operates at around 10 Frames per Second (FPS) for a video with frame size of $480 \times 720$ to as high as 56 FPS for a frame size of $240 \times 320$. This shows that even at an extremely low MR of 1\%, using our algorithm, effective and real-time tracking is possible by using CS measurements. 
				
				\vspace{-0.3cm}
				\section{Conclusion}
				\vspace{-0.1cm}
				In this paper we have presented a CNN-based non-iterative solution to the problem of CS image reconstruction. We showed that our algorithm provides high quality reconstructions on both simulated and real data for a wide range of measurement rates. Through a proof of concept real-time tracking application at the very low measurement rate of 0.01, we demonstrated the possibility of CS imaging becoming a resource-efficient solution in applications where the final goal is high-level image understanding rather than exact reconstruction. However, the existing CS imagers  are not capable of delivering real-time video. We hope that this work will give the much needed impetus to building of more practical and faster video CS imagers.

				\bibliographystyle{ieee}
				\bibliography{egpaper_final}
				
				\par\leavevmode
				
			\end{document}